\documentclass{bmvc2k}

\usepackage{amssymb}
\usepackage{amsmath}

\usepackage{algorithm}
\usepackage{algpseudocode}
\usepackage{makecell} % for thead.
\usepackage[normalem]{ulem} %Strikeout via sout
\usepackage[inline]{enumitem}

\usepackage[justification=centering]{caption}
\usepackage{graphicx}
\usepackage{subcaption}

\title{DCNNs: A Transfer Learning comparison of Full Weapon Family threat detection for Dual-Energy X-Ray Baggage Imagery}

\addauthor{Ashley Williamson}{ashley.williamson@hull.ac.uk}{1}
\addauthor{Patrick Dickinson}{pdickinson@lincoln.ac.uk}{2}
\addauthor{Tryphon Lambrou}{tlambrou@lincoln.ac.uk}{2}
\addauthor{John C. Murray}{john.murray@hull.ac.uk*}{1}

\addinstitution{
 Department of Computer Science and Technology\\
 University of Hull\\
 Hull, UK\\
 (* Corresponding Author)
}
\addinstitution{
 Department of Computer Science\\
 University of Lincoln\\
 Lincoln, UK
}

\runninghead{Williamson \etal}{Full-weapon X-Ray Threat Detection}

\def\etal{\emph{et al}\bmvaOneDot}

\newcommand{\imagenet}{ILSVRC }

\begin{document}

\maketitle

\begin{abstract}
%% Text of abstract
Recent advancements in Convolutional Neural Networks have yielded super-human levels of performance in image recognition tasks
\cite{DBLP:journals/corr/RussakovskyDSKSMHKKBBF14,he2015delving}; however, with increasing volumes of parcels crossing UK borders each year, classification of threats becomes integral to the smooth operation of UK borders.
In this work we propose the first pipeline to effectively process Dual-Energy X-Ray scanner output, and perform classification capable of distinguishing between firearm families (Assault Rifle, Revolver, Self-Loading Pistol, Shotgun, and Sub-Machine Gun) from this output.
With this pipeline we compare recent Convolutional Neural Network architectures against the X-Ray baggage domain via Transfer Learning and show ResNet50 to be most suitable to classification - outlining a number of considerations for operational success within the domain.
\end{abstract}
%-------------------------------------------------------------------------
\section{Introduction}
\label{sec:intro}
Dual-Energy X-Ray scanning systems are ubiquitous in border security applications, and pose a substantial challenge for automation - requiring trained officers for successful operation. These technologies are employed for a wide range of logistical solutions for passenger, commercial, industrial baggage and parcel services. With an ever increasing volume of parcels, systems are put under pressure to classify complex contents in shorter time-spans for detection of threats.\\
In recent years, significant advancements have been made in the field of Object Classification and Detection, specifically through the yearly ImageNet( \imagenet) competition \cite{DBLP:journals/corr/RussakovskyDSKSMHKKBBF14}. Whilst \imagenet is designed for general object classification, there has been little work applying such advancements specifically to the security domain.\\
Existing work towards Dual-Energy X-Ray baggage object detection focuses on traditional feature extraction, segmentation, enhancement, and detection algorithms to facilitate human operators in the interrogation of baggage imagery. Turcsany \textit{et al} \cite{turcsany2013improving} demonstrate a Visual Bag-of-Words model applied to 2D pseudo-colour images using DoG, DoG+SIFT, and DoG+Harris feature representations, with expansions \cite{bacstan2011visual} on such work focusing on the use of SURF \cite{bay2006surf} and SVM Classifiers - yielding improved classification results due to a large diverse dataset. In addition, Flitton \textit{et al} \cite{flitton2015object} propose 3D Computed Tomography (CT) imagery solutions extending on 2D methods via a combination of 3D Feature Descriptors - Density Histogram(DH), Density Gradient Histogram(DGH), SIFT, and Rotation Invariant Feature Transform(RIFT). Kechagias-Stamatis \textit{et al}\cite{kechagias2017automatic} outline a proposed pipeline relying on local feature extraction via SURF features, utilising soft and hard clustering. Further work has looked at enhancing image output as a means of improving object detection \cite{chen2005combinational}.
Ak{\c{c}}ay and Breckon \cite{akcay_2017} compare transfer learning within the domain of X-Ray Threat Detection on a limited-scope dataset comprised of disparate threats with various mechanisms such as Sliding Window CNN, and recent region proposal-based architectures concluding these approaches to be superior to hand-crafted features. Ak{\c{c}}ay \etal \cite{akcay_2018} continues this work - outlining datasets labelled $Dbp_{2}$ and $Dbp_{6}$ for firearm-not-firearm and mutli-class firearm/threat classification respectively - whereby classification and detection mechanisms are compared for both these datasets and classification is performed on Full-Firearm vs Operational Benign (FFOB) and Firearm Parts vs Operational Benign (FPOB); confirming application of Convolutional Neural Networks to outperform hand-crafted features. However \cite{akcay_2017, akcay_2018} include objects such as guns, knives, laptops as 'threat' objects when performing classification.
\\
Ak{\c{c}}ay \etal\cite{akccay2016transfer} compare the depth of representation freezing, when transfer learning, against accuracy with a pre-trained AlexNet\cite{AlexNet_NIPS2012_4824} model, showing benefits when freezing layers 1-3.
To the best of our knowledge, we are the first to consider various Deep Convolutional Neural Network models, including more recent models, for the application of transfer learning to this problem via a direct-from-scanner approach - where our dataset preprocessing enables us to produce classification directly from X-Ray Scanner Output, on a dataset constructed of 5 similar firearms of distinct families.

\subsection{Convolutional Neural Networks \& Transfer Learning}
Deep Convolutional Neural Networks have been applied to a host of domains since their inception, including Video classification \cite{karpathy2014large}, Reinforcement Learning \cite{mnih2013playing}, Natural Language Processing \cite{collobert2008unified}, and in recent years have surpassed human-level performance in image recognition tasks \cite{russakovsky2015imagenet, he2015delving}.\\
These networks provide a means of deeper image representation, where initial layers represent basic image features such as edges or boundaries, with further layers providing more abstract representations such as faces; dependent upon the training dataset \cite{zeiler2014visualizing}. 
%This is achieved via application of the convolution operation using a sliding window over an input image, producing feature maps. This operation is then repeated according to the given architecture sequentially upon each proceeding layer; where each new layer provides feature maps with progressively more abstraction - This is what constitutes deep networks. 
These representations are then combined with fully-connected layers to weight which features contribute towards the correct classification of a given class - often utilising softmax to provide class probability outputs.\\

Successful classification typically relies on substantial numbers of training examples to learn from, with \imagenet containing upwards of 14 millions images over 1000 classes - providing sufficient information to train CNNs from scratch. Evolution of Neural Network architectures are producing more accurate classification accuracies on \imagenet challenges, yet for domains where training examples are scarce, or expensive to obtain, training from scratch can be problematic or may lack sufficient data to adequately produce a model. Transfer Learning \cite{pan2010survey} exploits the innate ability of CNNs to produce feature abstraction, and applies this to a new domain not originally trained on, the \textit{target} domain. This technique has become popular across difficult training domains, and has been shown to work within detection scenarios \cite{ng2015deep, shin2016deep}. Transfer Learning involves taking the weights of a given architecture, trained to a high degree of accuracy on an \textit{existing} domain, and initialising a new model with those same weights for a different domain, the \textit{target}. 
%Convolutional Neural Networks are commonly discussed inclusive of their final fully-connected softmax layers, however in this context to successfully transfer a model one must reinterpret the convolutional layers outputs by discarding the original model's interpretation of the feature maps. The act of training a transferred model involves the creation of adaption layers beyond the convolutional layers, and typically consists of a sequence of fully-connected layers binding to new class labels - which may be of different shape - and is often comprised entirely of different classes from the \textit{existing} domain to the \textit{target} domain.
This approach significantly reduces training times by bootstrapping learning, and on occasion, prohibiting backpropagation into the earlier layers, focusing only on the final layers - fine-tuining. A variation upon this approach freezes a sub-set of the convolutional layers, enabling fine-tuning of the mid to high-level features \cite{oquab2014learning}.
Chollet \cite{chollet2016xception} states that training required 3 days on the original ILSVRC-2012 dataset, utilising 60 K80 GPUs; additionally Simonyan and Zisserman \cite{simonyan2014very} reported 3-4 weeks of training on NVidia Titan Black GPUs depending on the variant of their architecture used. With Transfer Learning we can re-use the knowledge of these original domains, and adapt them for Dual-Energy X-Ray Imagery within fractions of the time; when compared against training a CNN from random initialisation.
%-------------------------------------------------------------------------
\section{Experimental}
\subsection{Dataset}

\begin{figure}[t]
	\centering
  	\includegraphics[width=0.6\linewidth]{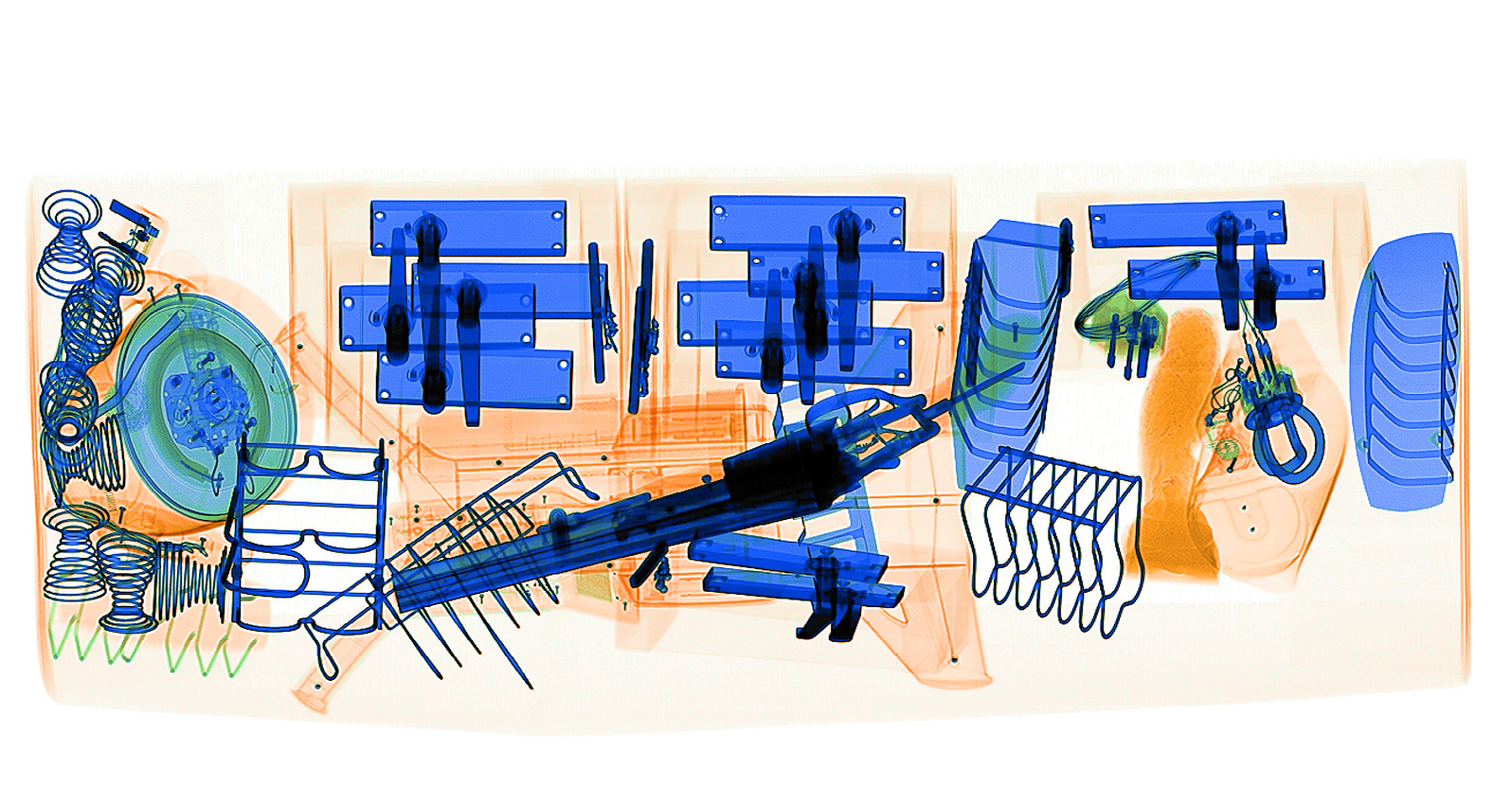}
  	\caption{Example of false colour representation of Dual-Energy X-Ray Imagery.}
    \label{fig:false-colour-baggage}
% Some examples of the false colour representation.
\end{figure}

% Provide example of dataset images.
% Imagegroups
We utilise a novel dataset provided by the Home Office's Centre for Applied Science and Technology (CAST), consisting of false-colour images of baggage items, where higher atomic weights are represented via blue hues, corresponding to metallics, and orange hues represent lower atomic weights, such as organic material; with greens being a mix of organic and in-organic materials (See Figure \ref{fig:false-colour-baggage}).
%Information on the exact colour mapping is unknown; however, these are standard mappings for Rapiscan equipment and are subject to slight variation given calibration.\\
Data is comprised of fullweapon examples only, and represents the following classes: \textit{assault rifle}, \textit{revolver}, \textit{self-loading pistol}, \textit{shotgun}, and \textit{sub-machine gun} with 2160 positive examples across all classes; containing 450, 450, 450, 360, and 450 examples per class, respectively. Each image belongs to an \textit{imagegroup}, where members of an \textit{imagegroup} correspond to the same physical baggage being scanned from multiple viewpoints; these include top-down, side-view, and $\pm$ 45 oblique, dependent upon manufacturer. These are split into training and testing example sets, whereby no imagegroup is bisected, with 70-30 ratio maintaining class distribution consistent across the set boundary. It is worth noting that no selective filtering is done upon the dataset to remove erroneous images, examples of which include distortion or empty images during image acquisition. Image labels were provided as-is from CAST via metadata related to each file. Final training set contains 1524 fullweapons, with 318, 318, 318, 252, 318 examples over respective classes; with the testing set containing 132, 132, 132, 108, 132 examples respectively.\\
%%% TODO: Reference Durham here with Dbp2 and Dbp6
Prior works have only sought to address a binary gun-not-gun problem, or a 6-class multi-object problem. Our dataset includes more difficult cases where differences between classes represent fundamental differences between specific gun families; whereby overlap of features will be commonplace. In addition, our dataset includes significantly fewer examples for this task.\\
To our knowledge, we are the first to consider sub-classes of firearm classification in this context, specifically in an end-to-end manner.

%\begin{minipage}{\linewidth}
\begin{algorithm}
	\begin{algorithmic}[1]
	\Require
		\Statex $function\ inRange(i, l, u)-$ produces a 0, 1 output if a given pixel lies between the lower-bound, l, and the upper-bound, u.
		\Statex
		\Statex $matrix\ J_{n} -$ unit matrix of $n$x$n$, composed of values 1.
		\Statex
		\Statex $function\ hsv(im_{bgr})-$ converting $im_{bgr}$ into HSV Colour Space.
		\Statex
		\Statex $function\ boundingRect(mask)-$ calculate minimum up-right bounding rectangle of non-zero elements of $mask$.
		\Statex		
		\Statex $function\ centroid(mask)-$ calculate the centroid of the given mask.
		\Statex		
		\Statex $function\ padd( image, top, bottom, left, right )-$ Pads the provided image with whitespace, by the amount specified in the given four directions.
	\Statex
	\State $B_{min} = (90,\ 100,\ 100)$
	\State $B_{max} = (180,\ 255,\ 255)$ 
	\State $images = \{im_{\text{0}},\ im_{\text{1}},\ \ldots,\ im_{\text{N}}\}$
    \State $c = \{c_{\text{0}},\ c_{\text{1}},\ \ldots,\ c_{\text{N}}\}$%\COMMENT Centroid array
	\State $meanWindow = [0,0]$
	\State $count = 0 $
	\For {$im_{hsv} \leftarrow hsv(im_{bgr}) \in images$}
		\State $mask_{hsv} \leftarrow inRange(im_{hsv}, B_{min}, B_{max})$
		\State $morphMask_{hsv} \leftarrow ( mask_{hsv} \ominus J_{3} ) \bullet J_{10}$
		\State $c_{im_{hsv}}\leftarrow centroid( morphMask_{hsv} )$
        \State $bRect\leftarrow boundingRect(morphMask_{hsv})$
			%MEAN_DIMS[0] += (1 / (count + 1)) * (w - MEAN_DIMS[0])
        	%MEAN_DIMS[1] += (1 / (count + 1)) * (h - MEAN_DIMS[1])
        \State $meanWindow \mathrel{{+}{=}} \frac{1}{ counter + 1 } \cdot ( bRect - meanWindow )$
        \State $counter = counter + 1$
	\EndFor
	\For {$im_{hsv} \leftarrow hsv(im_{bgr}), c_{im_{hsv}} \in images, c$}
		\State $bounds_{x} \leftarrow (c[0], c[0] + meanWindow[0])$
        \State $bounds_{y} \leftarrow (c[1], c[1] + meanWindow[1])$
		\State $padded_{hsv} \leftarrow padd($
          \State\hspace{\algorithmicindent} $image=im_{hsv},$
          \State\hspace{\algorithmicindent} $top= \lfloor \frac{meanWindow[1]}{2} \rfloor,$
          \State\hspace{\algorithmicindent} $bottom= \lceil \frac{meanWindow[1]}{2} \rceil,$
          \State\hspace{\algorithmicindent} $left= \lfloor \frac{meanWindow[0]}{2} \rfloor,$
          \State\hspace{\algorithmicindent} $right= \lceil \frac{meanWindow[0]}{2} \rceil$
        \State  $)$
		\State $final \leftarrow im_{hsv}[bounds_{x}[0]:bounds_{x}[1], bounds_{y}[0]:bounds_{y}[1]]$
		\State $save( resize( final, \frac{1}{2} ) )$
	\EndFor

	\caption{Maximal information bounding}
	\label{alg:max_info_window}
    \end{algorithmic}
\end{algorithm}
%\end{minipage}

\subsubsection{Preprocessing}

\begin{figure}[!h]
	\centering
    \begin{subfigure}[b]{.3\linewidth}
  		\frame{\includegraphics[width=\textwidth]{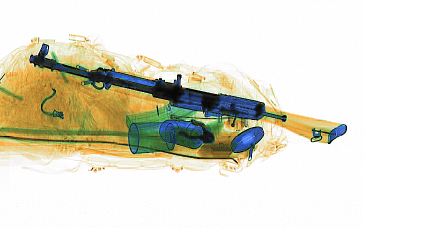}}
        \caption{Assault Rifle}
        \label{fig:training-examples-assault}
    \end{subfigure}
    \begin{subfigure}[b]{.3\textwidth}
  		\frame{\includegraphics[width=\textwidth]{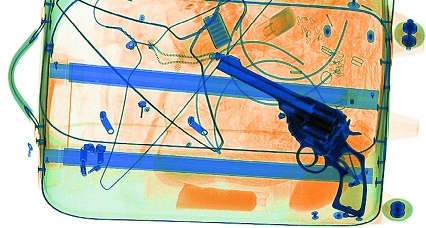}}
        \caption{Revolver}
        \label{fig:training-examples-revolver}
    \end{subfigure}
    \begin{subfigure}[b]{.3\textwidth}
  		\frame{\includegraphics[width=\textwidth]{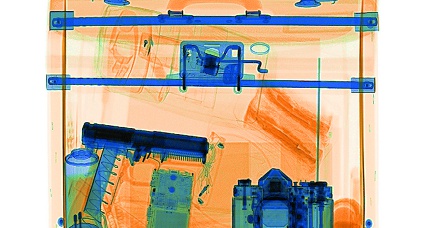}}
        \caption{Self-Loading Pistol}
        \label{fig:training-examples-pistol}
    \end{subfigure}
\end{figure}
\begin{figure}[!h]\ContinuedFloat\centering
    \begin{subfigure}[t]{0.3\textwidth}
  		\frame{\includegraphics[width=\textwidth]{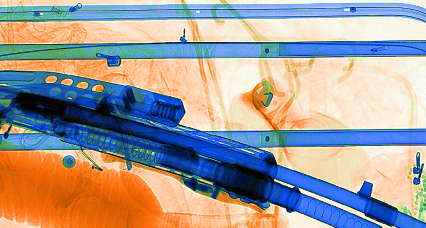}}
        \caption{Shotgun}
        \label{fig:training-examples-shotgun}
    \end{subfigure}
    \begin{subfigure}[t]{0.3\textwidth}
  		\frame{\includegraphics[width=\textwidth]{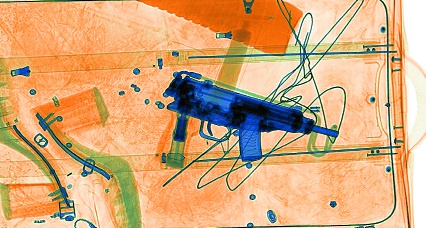}}
        \caption{Sub-Machine Gun}
        \label{fig:training-examples-smg}
    \end{subfigure}
    \vspace{2ex}
  	\caption{Training example images after maximal information windowing from 5 full-weapon categories. Prior to shorter-side cropping.}
    \label{fig:training-examples}
% Some examples of the false colour representation.
\end{figure}

Preprocessing consists of taking an output image from an X-Ray Scanner and processing it ready for interpretation by the Convolutional Neural Network. The same steps taken here apply for construction of the training dataset, as well as preprocessing of new images for inference only. Preliminary HSV slicing, between $(H=90,\ S=100,\ V=100)$ and $(H=180,\ S=255,\ V=255)$, to highlight high \textit{Effective Atomic Weight} (\textit{$Z_{eff}$}) values, is performed to segment metallic responses; positive threats within the dataset have high metallic components. Secondly, morphology operations reduce any smaller erroneous responses, as well as emphasise and focus on the primary cluster of high-response; representing the actual threat. From this we denote centroid locations, and bounding boxes of responses in order to calculate a \textit{mean response window}, for which our network will be shaped to. The intuition behind our approach is that high metallic responses will contain the maximal information from the sample, and thus creating a minimum bounding box around these responses will result in the highest likelihood of threat detection contributing to learning. This process is outlined in Algorithm \ref{alg:max_info_window}. As Convolutional Neural Networks containing fully-connected layers require a fixed input size, it is important to choose an appropriate input size; we chose the mean window response as an indication of aspect ratio - later resizing by $\frac{1}{2}$ to reduce memory usage and complexity for processing. Examples of preprocessing output can be seen in Figure \ref{fig:training-examples}.\\

As data provided consists of Multi-View Dual-Energy X-Ray images of baggage, it is important to ensure that those images which represent the same physical specimen be grouped such that they entirely lie within either the training set, or the test set; due to high similarity between images of the same image group. Therefore we employ an image group split mechanism as a means of ensuring our training-test split is as close to ideal as possible - 70-30 training-testing split. We maintain class balance over the sets via this process, such that the distribution amongst classes pre-split is as close as possible to the post-split, whilst still adhering to imagegroup boundaries.\\
After splitting, the training set contains 1524 fullweapons, with 318, 318, 318, 252, 318 examples over respective classes; with the testing set containing 132, 132, 132, 108, 132 examples respectively. To utilise this dataset with the original networks we perform shorter-side cropping to the two modes of input dimension, 224x224, or 299x299, when feeding the network.
\subsection{Framework}
To enable a direct comparison, an evaluative framework was developed which encapsulates each specific network, acting as an interface for standard training/testing operations. These include \textit{building}, \textit{training}, \textit{testing}, \textit{loading}, and \textit{saving} each network. Tensorflow \cite{tensorflow2015-whitepaper}, and Keras \cite{chollet2015keras} were used to realise this framework, with Keras providing a substantial number of the models with existing weights trained from the \imagenet domain. AlexNet \cite{AlexNet_NIPS2012_4824} was originally under the Caffe framework \cite{jia2014caffe}, with the architecture obtained from Tensorflow/Models Github \cite{tensorflow/models/alexnet} for Tensorflow with a conversion of the original weights being provided by Michael Guerzhoy \cite{bvlc_alexnet.npy}. All other pre-trained weights were provided via Keras implementations.\\
Models selected for comparison include AlexNet \cite{AlexNet_NIPS2012_4824}, VGG19 \cite{simonyan2014very}, ResNet50 \cite{he2016deep}, InceptionV3 \cite{szegedy2016rethinking}, and Xception \cite{chollet2016xception}. We use colloquial nomenclature to enable reproducibility and linking between implementation and theory; where VGG19 is equivalent to VGG Model D, and ResNet50 is a Residual Network of length 50.

\subsection{Training}
Each model is built following the architecture outlined by their respective implementations, whereby we perform shorter-side cropping of either 224x224 or 299x299, centrally resizing to the target dimensions. We re-implement a standard \textit{top-layer} on-top of each convolutional neural network for the given classification task, consisting of ReLU \cite{relu} activation functions, terminated by a softmax output.
%Each Convolutional Neural Network model evaluated has specific input dimensions, as a result the dataset after preprocessing must undergo final modifications to fit the network. Input falls into two categories: 224x224 or 299x299. We perform shorter-side cropping, then centrally resize to the target dimensions.\\
%Each model is built following the architecture outlined in their respective implementations, in order to transfer these networks we re-implement the top-layers of the network - The fully-connected layers ontop of the convolutional layers. This varies based on the model, and typically involves some \textit{fully-connected} layers, with ReLU \cite{relu} activation functions, terminated in a softmax wrapped output. As we are applying this model to our new domain the softmax output differs from original implementations, as we are only dealing with 5 classes. In order to prevent modification of the image representation, 
We apply a stop mechanism between the convolutional layers and the redefined \textit{top-layers} preventing any gradient calculation being propagated backwards and modifying the weights of the earlier layers of the networks; facilitating faster learning by reducing the number of trainable parameters calculated.\\

\label{label:stopping}
From the Model definition we use a Stochastic Gradient Descent Optimiser with $lr = 1^{-3}$, $momentum = 0.9$, and $decay = 1^{-4}$, with $batch size = 64$ for all models. Batching is done by randomly sampling from the given set, without replacement. Each epoch represents a processing of all batches from the dataset. AlexNet model parameters \cite{bvlc_alexnet.npy} are loaded via TensorFlow, with top-layer weights and biases being initialised via truncated normal distribution with $\mu=0.0$, and $\sigma=0.001$. Remaining models are initialised using ImageNet weights provided by Keras for Convolutional Layers, with custom top-layer weights being randomly initialised via \textit{glorot uniform distribution}(Xavier uniform distribution), and zero initialised bias units - as default.\\
Whilst training we use early stopping, such that if $k$ consecutive epochs loss value does not improve (minimise) we halt training and return the model with the lowest loss. We denote an $upperlimit=3000$ as our absolute upper-bound on number of epochs to train, and use $k=50$ for stopping.\\
Each model is trained on dual Intel(R) Xeon(R) CPU E5-2620 v4 @ 2.10GHz CPUs, 8x16GB Samsung DDR4 Registered DIMMs @ 2667 MT/s, with a single NVidia Titan XP GPU with 3840 CUDA cores, running TensorFlow 1.4.0-rc1 compiled from source. Models were trained in parallel, each with their own dedicated card; however sharing system resources for CPU and RAM.

%% Theory/calculation
%% A Theory section should extend, not repeat, the background to the article already dealt with in the Introduction
%% and lay the foundation for further work. In contrast, a Calculation section represents a practical development from a theoretical basis.
%\section{Theory/Calculation}

%% Discussion
%% This should explore the significance of the results of the work, not repeat them. A combined Results and Discussion section is often appropriate. Avoid extensive citations and discussion of published literature.

\section{Results \& Discussion}

Average Inference time per image was calculated based on 500 iterations of the test-set, obtaining the time over all epochs, averaging over this sum, followed by division of number of test samples within an epoch to obtain the average image response time. For Sensitivity and Specificity calculations, these were conducted on a one-vs-all approach for each model, calculated from the generated confusion matrices of each model.\\
Threat Detection algorithms do not work alone, and are typically part of a larger system; This, in combination with an increasing volume of parcels and baggage being processed by X-Ray scanning equipment, places a large emphasis on minimising processing time whilst maintaining accuracy for successful operation.
In order to evaluate the usefulness of each network tested, for the domain of Threat Detection for Dual-Energy X-Ray systems, we propose the consideration of the following criteria:
\begin{enumerate*}[label={\alph*) },font={\bfseries}]
\item retrainability,
\item high accuracy,
\item reduced parameters, and
\item low inference.
\end{enumerate*}\\
The ability of deep learning to learn complex visual problems combined with a reduced time-response to retraining are advantageous in a domain where the threat landscape is ever-changing.
The system must be robust to these introductions, and be able to quickly be redeployed promptly following identification and acquisition of new threat information.

Detection of threats at border control has a direct impact upon the safety of the population, the ability for the approach to classify weapons to a high accuracy is important and should be considered safety critical; with misclassification or omission resulting in severe consequences.

Reduced parameter count enables more images to be processed simultaneously by a single GPU, prompting larger scan volumes due to reduced number of operations required. Fewer parameters by the model need to be stored on the GPU, more room is freed up to dedicate to data processing. In addition, fewer trainable parameters directly influences the time taken to train the model sufficiently, as fewer gradients need to be calculated in a single backpropagation pass. If the model can be initialised and ran utilising less GPU memory, it directly results in cheaper implementation costs; using existing consumer-grade hardware within scanning equipment.\\
As threat detection solutions do not operate in isolation but in tandem, low inference times are essential to ensure that the impact of the threat detection pipeline as a whole is not impeded; if classification cannot be performed in a timely manner this can cause reduction in throughput of border control and distribution centres, and overall disruption.\\

\begin{figure}[]
\centering
\begin{subfigure}{0.8\linewidth}
  \centering
  \includegraphics[width=0.8\linewidth]{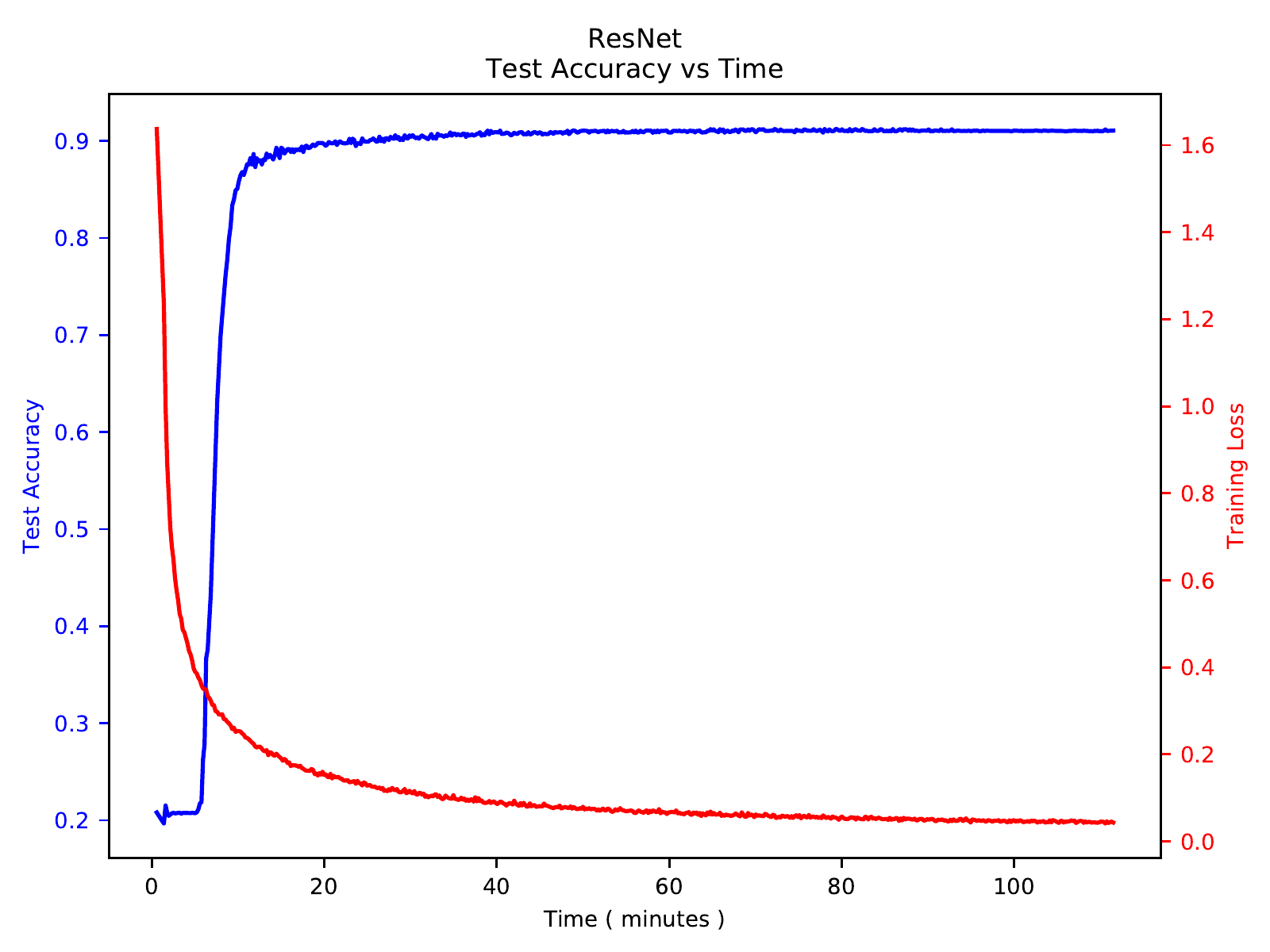}
\end{subfigure}
\caption{ResNet50 Training Accuracy/Loss vs Time(minutes)}
\label{fig:resnet-loss}
\end{figure}

From the overall results (See table \ref{tbl:network-training-inference}) it can be seen that newer architectures have a trend towards fewer parameters with the most recent, Xception, leading in this category. The architecturally simpler networks of AlexNet, and VGG19 lend themselves to lower training times, due in-part to their low inference times allowing higher throughput.
%Initial stopping metrics were explored, namely PQ Early-Stopping \cite{prechelt2012early}, however this yielded all networks running to absolute epoch length of 3000, requiring the utilisation of a different stopping mechanism. Therefore, we utilised a simpler mechanism, based on consecutive non-improvement, outlined previously in the training section. 
We found consecutive stopping criteria to be the most effective when applying transfer learning, as the loss function was relatively smooth - PQ Early-Stopping \cite{prechelt2012early} was designed with more noisy functions in mind, and was therefore not beneficial in our scenario and thus discarded. With our previously defined stopping criteria ( See section \ref{label:stopping} ) mechanism we achieve a best training of 26.3 minutes for VGG19, with Xception taking 814.9 minutes of training. Whilst the overall stopping time for ResNet50 is denoted as 111.47 minutes (See table \ref{tbl:network-training-inference}), the test-set accuracy plateaus relatively quickly (See figure \ref{fig:resnet-loss}) showing the reported training time as an upper-bound, where highest-accuracy models, are saved and output significantly earlier in the training process. With reference to Figure \ref{fig:resnet-loss}, training time for ResNet50 can be shown to be comparable to VGG19, with both models having similar inference times of 4.7 and 4.2 respectively. Of models tested, AlexNet yields the lowest accuracy of 77.51\% with the latest models, InceptionV3 and Xception, performing with 81.13\%, and 84.43\% respectively. Surprisingly the larger, more simple, VGG19 network out-performs these within this domain with 88.68\%. Overall ResNet50 produces the highest test-set accuracy of 91.04\%, a 2.36\% improvement over VGG19.
Of these networks both VGG19 and ResNet50 boast a low BER per-class with a low of 5.01\% and 3.35\% respectively; other models produced BER typically between 10 - 20\%. Further metrics from each model can be seen in tables \ref{tbl:alexnet-sens-spec}, \ref{tbl:vgg19-sens-spec}, \ref{tbl:resnet50-sens-spec}, \ref{tbl:inceptionv3-sens-spec}, and \ref{tbl:xception-sens-spec} - with reference to table \ref{weapon-lookup} for a reference key for the class id.

\begin{table}[!ht]
\tiny
%\captionsetup{justification=centering}
\caption{CNN architectures with Parameters, Training Times (hours), and Average Inference Times (ms) over 500 test-set runs, and test-set accuracy.}
\vspace{2ex}
\label{tbl:network-training-inference}
\centering
\resizebox{\linewidth}{!}{%
\begin{tabular}{l|c|c|c|c}
\multicolumn{1}{c|}{Model Name}
%& {\thead{\tiny Number of\\ \tiny Parameters}} & {\thead{\tiny Transfer\\\tiny Training\\ \tiny Time (minutes)}} & {\thead{\tiny Average %Inference\\\tiny Time Per Image (ms)}} & \thead{\tiny Test-set\\\tiny Accuracy(\%)}\\ 
& {Number of Parameters} & {Transfer Training Time (minutes)} & {Average Inference Time Per Image (ms)} & {Test-set Accuracy (\%)}\\
\hline
AlexNet \cite{AlexNet_NIPS2012_4824} &  111,443,342 &  70.40 &  \textbf{1.35} & 77.51 \\
VGG19 \cite{simonyan2014very} & 55,704,649 & \textbf{26.3} &  4.70 & 88.68\\
Resnet50 \cite{he2016deep} & 23,597,961 & 111.47 &  4.2 & \textbf{91.04}\\
InceptionV3 \cite{szegedy2016rethinking} & 21,813,033 & 370.1 &  6.27 & 81.13\\
Xception \cite{chollet2016xception} & \textbf{20,871,729} & 814.9 & 8.54 & 84.43
\end{tabular}%
}
\end{table}

%\begin{table}
%%\captionsetup{justification=centering}
%\caption{Lookup table mapping Class ID to Full Weapon Category}
%\vspace{2ex}
%\label{weapon-lookup}
%\centering
%\begin{tabular}{l|c}
%\multicolumn{1}{c|}{Model Name} & {Class ID}\\
%\hline
%Assault Rifle 		& 0\\
%Revolver 			& 1\\
%Self-Loading Pistol & 2\\
%Shotgun 			& 3\\
%Sub-Machine Gun 	& 4
%\end{tabular}
%\end{table}

\begin{table}
\tiny
\vspace*{-\baselineskip}
%\captionsetup{justification=centering}
\caption{Lookup table mapping Class ID to Full Weapon Category}
\vspace{2ex}
\label{weapon-lookup}
\centering
\begin{tabular}{l|p{.1\linewidth}|p{.1\linewidth}|p{.15\linewidth}|p{.1\linewidth}|p{.15\linewidth}}
\multicolumn{1}{c|}{Class ID} & {0} & {1} & {2} & {3} & {4}\\
\hline
Category &
Assault \newline Rifle &
Revolver &
Self-Loading \newline Pistol &
Shotgun &
Sub-Machine \newline Gun
\end{tabular}
\end{table}

%%% ALEXNET
\vspace*{-\baselineskip}
\begin{table}[!ht]
\tiny
%\captionsetup{justification=centering}
\caption{AlexNet per class classification metrics - each class is treated as a one-vs-all approach.}
\vspace{2ex}
\label{tbl:alexnet-sens-spec}
\centering
\resizebox{\linewidth}{!}{%
\begin{tabular}{c|c|c|c|c|c|c|c|c}
\multicolumn{1}{c|}{Class} & TP & TN & FP        & \multicolumn{1}{l|}{FN} & \multicolumn{1}{l|}{Sens (\%)} & \multicolumn{1}{l|}{Spec (\%)} & \multicolumn{1}{l}{Acc (\%)} & BER(\%) \\ \hline
0 & 110 & 470 & 34 & 22 & 83.33 & 93.25 & 91.20 & 11.71\\
1 & 90  & 462 & 42 & 42 & 68.18 & 91.67 & 86.79 & 20.08\\
2 & 97  & 489 & 15 & 35 & 73.48 & 97.02 & 92.13 & 14.75\\
3 & 86  & 516 & 12 & 22 & 79.62 & 97.72 & \textbf{94.65} & \textbf{11.32}\\
4 & 110 & 464 & 40 & 22 & 83.33 & 92.06 & 90.26 & 12.30
\end{tabular}%
}
\end{table}

%%% VGG19
\vspace*{-\baselineskip}
\begin{table}[!ht]
\tiny
%\captionsetup{justification=centering}
\caption{VGG19 per class classification metrics - each class is treated as a one-vs-all approach.}
\vspace{2ex}
\label{tbl:vgg19-sens-spec}
\centering
\resizebox{\linewidth}{!}{%
\begin{tabular}{c|c|c|c|c|c|c|c|c}
\multicolumn{1}{c|}{Class} & TP & TN & FP        & \multicolumn{1}{l|}{FN} & \multicolumn{1}{l|}{Sens (\%)} & \multicolumn{1}{l|}{Spec (\%)} & \multicolumn{1}{l}{Acc (\%)} & BER(\%) \\ \hline
0 & 122 & 491 & 13 & 10 & 92.42 & 97.42 & \textbf{96.38} & 5.08\\
1 & 113 & 496 & 8  & 19 & 85.60 & 98.41 & 95.75 & 7.99\\
2 & 109 & 497 & 7  & 23 & 82.58 & 98.61 & 95.28 & 9.41\\
3 & 96  & 504 & 24 & 12 & 88.89 & 95.45 & 94.33 & 7.83\\
4 & 124 & 484 & 20 & 8  & 93.94 & 96.03 & 95.60 & \textbf{5.01}
\end{tabular}%
}
\end{table}

%%% ResNet50
\vspace*{-\baselineskip}
\begin{table}[!ht]
\tiny
%\captionsetup{justification=centering}
\caption{ResNet50 per class classification metrics - each class is treated as a one-vs-all approach.}
\vspace{2ex}
\label{tbl:resnet50-sens-spec}
\centering
\resizebox{\linewidth}{!}{%
\begin{tabular}{c|c|c|c|c|c|c|c|c}
\multicolumn{1}{c|}{Class} & TP & TN & FP        & \multicolumn{1}{l|}{FN} & \multicolumn{1}{l|}{Sens (\%)} & \multicolumn{1}{l|}{Spec (\%)} & \multicolumn{1}{l}{Acc (\%)} & BER(\%) \\ \hline
0 & 125 & 497 & 7  & 7  & 94.70 & 98.61 & 97.80 & \textbf{3.35}\\
1 & 109 & 492 & 12 & 23 & 82.58 & 97.62 & 94.50 & 9.90\\
2 & 121 & 488 & 16 & 11 & 91.67 & 96.83 & 95.75 & 5.75\\
3 & 102 & 521 & 7  & 6  & 94.44 & 98.67 & \textbf{97.96} & 3.44\\
4 & 122 & 489 & 15 & 10 & 92.42 & 97.02 & 96.07 & 5.28
\end{tabular}%
}
\end{table}

%%% InceptionV3
\vspace*{-\baselineskip}
\begin{table}[!ht]
\tiny
%\captionsetup{justification=centering}
\caption{InceptionV3 per class classification metrics - each class is treated as a one-vs-all approach.}
\vspace{2ex}
\label{tbl:inceptionv3-sens-spec}
\centering
\resizebox{\linewidth}{!}{%
\begin{tabular}{c|c|c|c|c|c|c|c|c}
\multicolumn{1}{c|}{Class} & TP & TN & FP        & \multicolumn{1}{l|}{FN} & \multicolumn{1}{l|}{Sens (\%)} & \multicolumn{1}{l|}{Spec (\%)} & \multicolumn{1}{l}{Acc (\%)} & BER(\%) \\ \hline
0 & 118 & 488 & 16 & 14 & 89.39 & 96.83 & 95.28 & \textbf{6.89}\\
1 & 96  & 481 & 23 & 36 & 72.73 & 95.44 & 90.72 & 15.92\\
2 & 100 & 479 & 25 & 32 & 75.76 & 95.04 & 91.04 & 14.60\\
3 & 95  & 513 & 15 & 13 & 87.96 & 97.16 & \textbf{95.60} & 7.44\\
4 & 107 & 463 & 41 & 25 & 81.06 & 91.87 & 89.62 & 13.54
\end{tabular}%
}

\end{table}

%%% Xception
\vspace*{-\baselineskip}
\begin{table}[!ht]
\tiny
%\captionsetup{justification=centering}
\caption{Xception per class classification metrics - each class is treated as a one-vs-all approach.}
\vspace{2ex}
\label{tbl:xception-sens-spec}
\centering
\resizebox{\linewidth}{!}{%
\begin{tabular}{c|c|c|c|c|c|c|c|c}
\multicolumn{1}{c|}{Class} & TP & TN & FP        & \multicolumn{1}{l|}{FN} & \multicolumn{1}{l|}{Sens (\%)} & \multicolumn{1}{l|}{Spec (\%)} & \multicolumn{1}{l}{Acc (\%)} & BER(\%) \\ \hline
0 & 119 & 484 & 20 & 13 & 90.15 & 96.03 & 94.81 & \textbf{6.91}\\
1 & 108 & 489 & 15 & 24 & 81.82 & 97.02 & 93.87 & 10.58\\
2 & 105 & 481 & 23 & 27 & 79.55 & 95.44 & 92.14 & 12.51\\
3 & 94  & 516 & 12 & 14 & 87.04 & 97.73 & \textbf{95.91} & 7.62\\
4 & 111 & 475 & 29 & 21 & 84.09 & 94.24 & 92.14 & 10.83
\end{tabular}%
}
\end{table}

% ResNet

\clearpage % make sure biblio is last.
\bibliography{bmvc_final}
\end{document}